\documentclass[conference]{IEEEtran}
\IEEEoverridecommandlockouts
\usepackage{multirow}
\usepackage{colortbl}
\usepackage{cite}
\usepackage{amsmath,amssymb,amsfonts}
\usepackage{algorithmic}
\usepackage{graphicx}
\usepackage{textcomp}
\usepackage{xcolor}
\usepackage{hyperref}
\def\BibTeX{{\rm B\kern-.05em{\sc i\kern-.025em b}\kern-.08em
    T\kern-.1667em\lower.7ex\hbox{E}\kern-.125emX}}
\begin{document}

\title{ASIGN: An Anatomy-aware Spatial Imputation Graphic Network for 3D Spatial Transcriptomics
}

\author{\IEEEauthorblockN{Junchao Zhu}
\IEEEauthorblockA{\textit{Department of Computer Science} \\
\textit{Vanderbilt University}\\
% City, Country \\
% email address or ORCID
}
\and
\IEEEauthorblockN{Ruining Deng}
\IEEEauthorblockA{\textit{Department of Computer Science} \\
\textit{Vanderbilt University}}

\and
\IEEEauthorblockN{Tianyuan Yao}
\IEEEauthorblockA{\textit{Department of Computer Science} \\
\textit{Vanderbilt University}}

\and
\IEEEauthorblockN{Juming Xiong}
\IEEEauthorblockA{\textit{Department of Electrical and Computer Engineering} \\
\textit{Vanderbilt University}}

\and
\IEEEauthorblockN{Chongyu Qu}
\IEEEauthorblockA{\textit{Department of Electrical and Computer Engineering} \\
\textit{Vanderbilt University}}

\and
\IEEEauthorblockN{Junlin Guo}
\IEEEauthorblockA{\textit{Department of Electrical and Computer Engineering} \\
\textit{Vanderbilt University}}

\and
\IEEEauthorblockN{Siqi Lu}
\IEEEauthorblockA{\textit{Department of Electrical and Computer Engineering} \\
\textit{Vanderbilt University}}

\and
\IEEEauthorblockN{Mengmeng Yin}
\IEEEauthorblockA{\textit{Department of Pathology, Microbiology and Immunology} \\
\textit{Vanderbilt University Medical Center}}

\and
\IEEEauthorblockN{Yu Wang}
\IEEEauthorblockA{\textit{Department of Biostatistics} \\
\textit{Vanderbilt University Medical Center}}

\and
\IEEEauthorblockN{Shilin Zhao}
\IEEEauthorblockA{\textit{Department of Biostatistics} \\
\textit{Vanderbilt University Medical Center}}

\and
\IEEEauthorblockN{Haichun Yang}
\IEEEauthorblockA{\textit{Department of Pathology, Microbiology and Immunology} \\
\textit{Vanderbilt University Medical Center}}

\and
\IEEEauthorblockN{Yuankai Huo}
\IEEEauthorblockA{\textit{Department of Computer Science} \\
\textit{Vanderbilt Unversity}\\
yuankai.huo@vanderbilt.edu}

}

\maketitle
\begin{abstract}
Spatial transcriptomics (ST) is an emerging technology that enables medical computer vision scientists to automatically interpret the molecular profiles underlying morphological features. Currently, however, most deep learning-based ST analyses are limited to two-dimensional (2D) sections, which can introduce diagnostic errors due to the heterogeneity of pathological tissues across 3D sections. Expanding ST to three-dimensional (3D) volumes is challenging due to the prohibitive costs; a 2D ST acquisition already costs over 50 times more than whole slide imaging (WSI), and a full 3D volume with 10 sections can be an order of magnitude more expensive. To reduce costs, scientists have attempted to predict ST data directly from WSI without performing actual ST acquisition. However, these methods typically yield unsatisfying results. To address this, we introduce a novel problem setting: 3D ST imputation using 3D WSI histology sections combined with a single 2D ST slide. To do so, we present the Anatomy-aware Spatial Imputation Graph Network (ASIGN) for more precise, yet affordable, 3D ST modeling. The ASIGN architecture extends existing 2D spatial relationships into 3D by leveraging cross-layer overlap and similarity-based expansion. Moreover, a multi-level spatial attention graph network integrates features comprehensively across different data sources. We evaluated ASIGN on three public spatial transcriptomics datasets, with experimental results demonstrating that ASIGN achieves state-of-the-art performance on both 2D and 3D scenarios. Code is available at \url{https://github.com/hrlblab/ASIGN}.

\end{abstract}  


\section{Introduction}
\label{sec:intro}

% Introduction for ST and data cost issue
The rapid advancement of spatial transcriptomic (ST) has provided researchers with an unprecedented capability to characterize spatial gene expression patterns at both the tissue and cellular levels \cite{burgess2019spatialcscd, asp2020spatially}, transforming our understanding of human tissue development, disease progression, and clinical diagnosis \cite{he2020integrating, maniatis2019spatiotemporal, asp2019spatiotemporal}.  
However, current spatial transcriptomics methods \cite{wang2018three, eng2019transcriptome, staahl2016visualization} are laborious and costly. This, coupled with the need for specialized equipment and expertise, limits their practical applications in real clinical settings.

\begin{figure}[t]
    \centering
    \includegraphics[width=\linewidth]{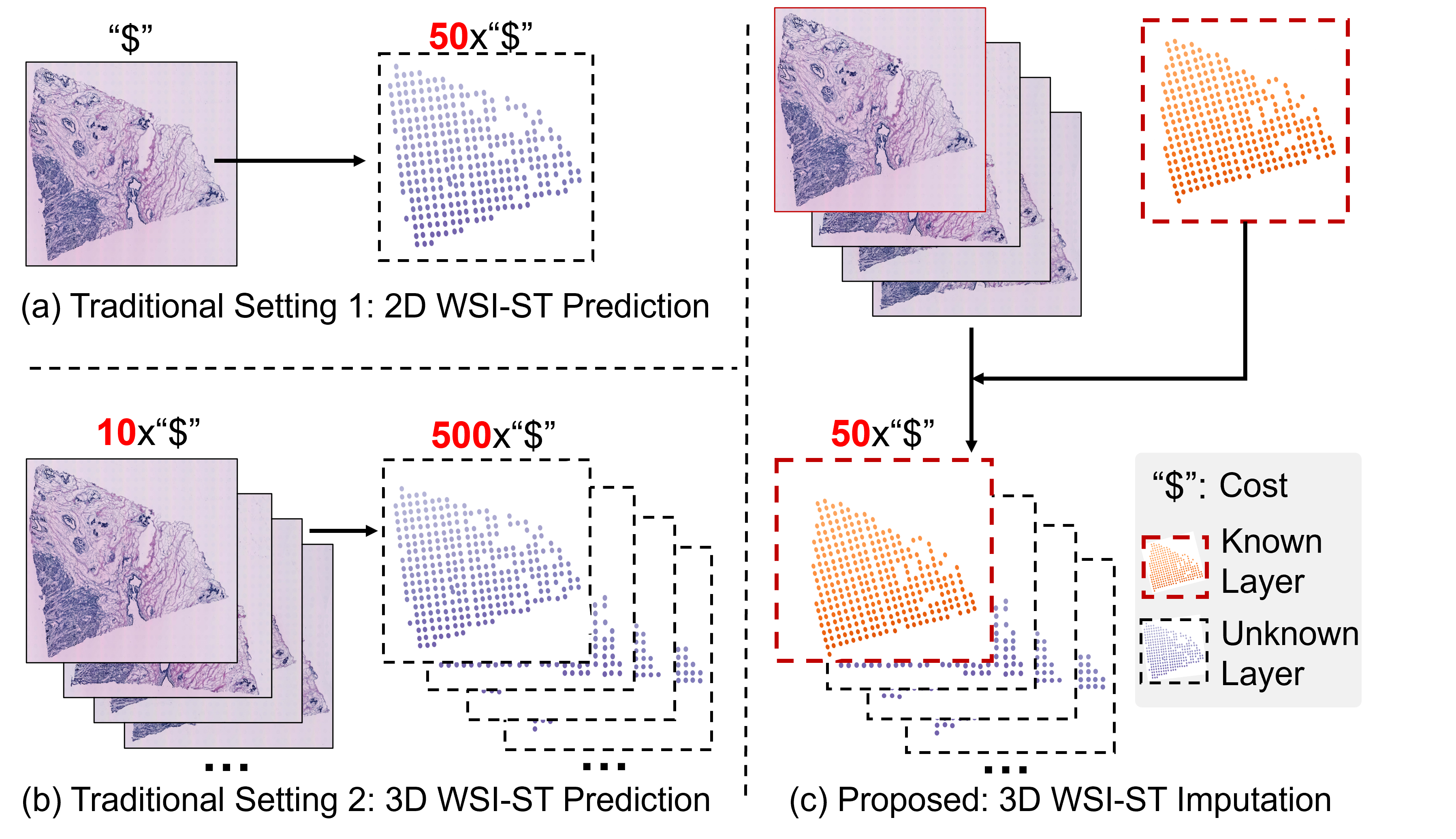}
    \caption{\textbf{ASIGN presents a novel problem setting for 3D spatial transcriptomic (ST).} Unlike recent methods that focus on either (a) directly predicting 2D ST from 2D WSI or (b) performing a full 3D WSI to 3D ST prediction, our proposed ASIGN method generates a 3D ST volume by combining 3D WSI histology sections with a single 2D ST slide. This approach provides a balanced solution—more precise than ``free but less accurate'' 3D predictions, yet far more affordable than the ``precise but prohibitively expensive'' acquisition of full 3D ST for every slide.}
    \label{fig:problem_definition}
\end{figure}

% Deep learning in pathology and ST
The features of histopathological images are closely correlated with gene expression patterns \cite{ash2021joint}, which offers an opportunity to directly predict spatial transcriptomic expression from images. 
Excitingly, several studies have employed architectures such as convolutional neural networks (CNNs) \cite{he2020integrating, yang2023exemplar, xie2024spatially, chung2024accurate} and graph neural networks (GNNs) \cite{pang2021leveraging, jia2024thitogene, zeng2022spatial} to predict spatial transcriptomic expression from histopathological two-dimensional whole-slide images (WSIs). 
These methods incorporate spatial dependencies \cite{xie2024spatially} and image similarities \cite{pang2021leveraging}, enabling predictions at the WSI level. This approach provides new pathways for analyzing single-layer tissue sections.

% Limitations: 3D information and cross-sample variance
However, there are still many daunting limitations in existing methods and training strategies. 
First, current approaches only rely on two-dimensional (2D) views of single-layer sections for spatial transcriptomic analysis and training, which neglects the rich structural information of organs and tissues in three-dimensional (3D) space. It might lead to diagnostic errors due to the heterogeneity of pathological tissues across 3D sections. Moreover, existing models exhibit noticeable deficiencies in generalizability, especially at the WSI level. This is mainly due to significant variance in gene expression across patients \cite{taylor2024sources, wolf2023characterizing}. Additionally, variations in staining styles \cite{runz2021normalization, bentaieb2017adversarial} and technical discrepancies in sequencing methods \cite{zhao2024innovative} across different data sources further contribute to this issue. 
Consequently, the performance of these models is often degraded when applied to new samples.

% Our methods
In this paper, We introduce an Anatomy-aware Spatial Imputation Graph Network (ASIGN) tailored for 3D ST analysis within a novel problem framework as is described in Figure \ref{fig:problem_definition}. Unlike recent approaches that either predict 2D ST directly from 2D WSI or perform full 3D WSI to 3D ST prediction, ASIGN creates the 3D ST volume by integrating 3D WSI histology sections with a single 2D ST slide. This method strikes a balance between ``free but less accurate" 3D predictions and the ``precise yet prohibitively costly" acquisition of full 3D ST on a slide-by-slide basis. ASIGN provides a new cost-effective route towards 3D ST, achieving a significant cost reduction while maintaining robust performance. 

To do so, ASIGN employs multiple registration methods to establish inter-layer overlaps at the 3D sample level, effectively constructing spot connections across samples. It also integrates a Multi-Level Spatial Attention Graph Network (MSAGNet) to enhance information fusion across layers, neighboring regions, and resolutions, enabling comprehensive feature integration. Additionally, ASIGN shifts training and prediction from an entirely unknown 2D WSI level to a partially known 3D sample level. To improve the imputation performance, we introduce a Cross-layer Imputation (CLI) block that propagates labels from known to unknown layers, enhancing the model's generalizability on unseen tissue samples.

% Contributions
Our contributions can be summarized as three folds:
\begin{itemize}
% 3D information
\item We introduce a new learning paradigm that shifts from 2D WSI-ST predictions to partially known 3D volumetric WSI-ST imputation. This is facilitated by a Cross-layer Imputation (CLI) block, which propagates information across layers to achieve more precise, yet affordable, 3D ST modeling.

\item We present ASIGN, a novel framework for 3D ST prediction that transforms isolated 2D spatial relationships into a cohesive 3D structure through inter-layer overlap and similarity, integrating 3D spatial information into 3D ST imputation.

% Network structure
\item A new multi-level spatial attention graphic network is proposed to facilitate comprehensive feature integration across different layers, neighboring regions, and multiple resolutions, thus enabling precise sample-level predictions.

\end{itemize}

\begin{figure}[htb]
    \centering
    \includegraphics[width=\linewidth]{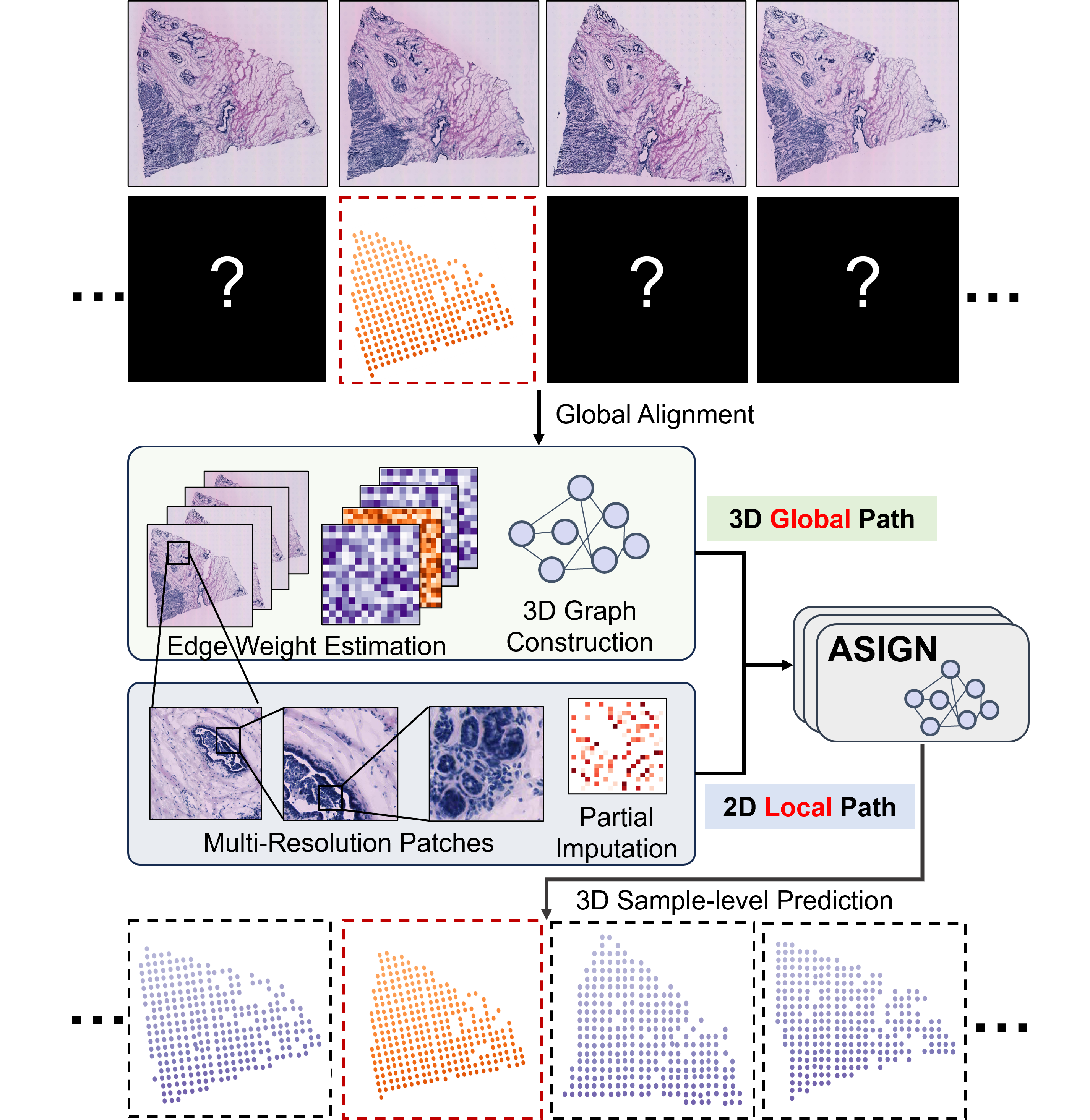}
    \caption{\textbf{Overall framework of our proposed ASIGN approach.}  ASIGN begins with a global alignment process to estimate overlap and similarity across layers, constructing 3D graph connections. Then, MSAGNet, integrated within ASIGN, enhances information fusion across layers, neighboring regions, and resolutions for comprehensive feature integration. Finally, a CLI block propagates predictions from partially known labels, which are weighted with MSAGNet’s model predictions to generate the final output.}
    \label{fig:Overall_framework}
\end{figure}

\begin{figure*}[htb]
    \centering
    \includegraphics[width=1\linewidth]{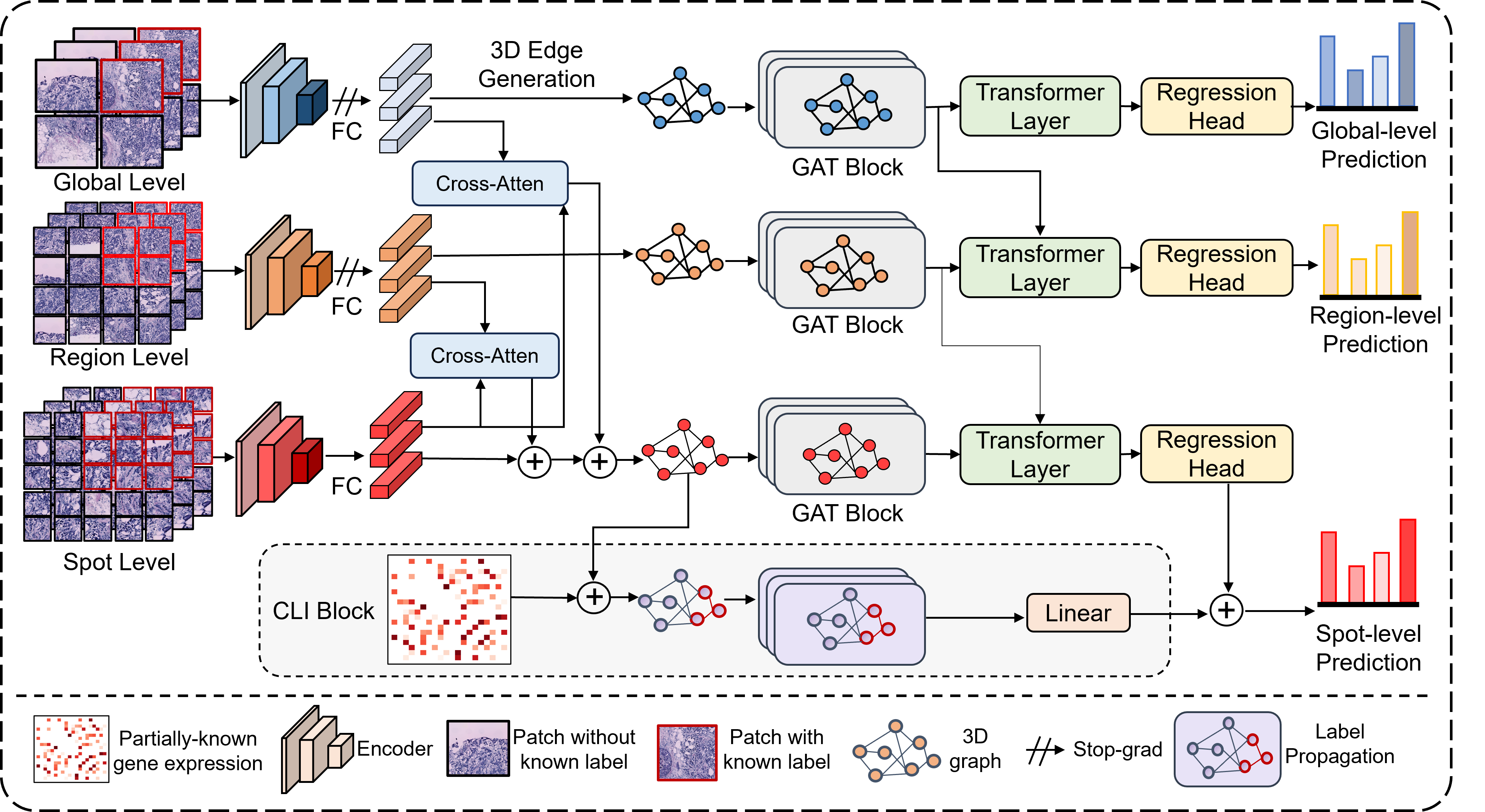}
    \caption{\textbf{Architecture of MSAGNet and the CLI block.} MSAGNet comprises cross-attention layers, GAT blocks, and Transformer layers to integrate and aggregate features across multiple resolution levels, 3D sample levels, and patch-self levels, respectively. The CLI block employs a label propagation strategy to diffuse known labels to unknown patches. A self-adaptive weighting mechanism merges the final results from both MSAGNet and the CLI block.}
    \label{fig:network_structure}
\end{figure*}

\section{Related Work}
\label{sec:formatting}

\subsection{Deep Learning Approaches for ST Prediction}
In recent years, deep learning methods have led to the emergence of various approaches for spatial gene expression prediction \cite{he2020integrating, yang2023exemplar, xie2024spatially,chung2024accurate,pang2021leveraging,zeng2022spatial}. ST-Net \cite{he2020integrating} is the earliest proposed method. It employs a transfer learning strategy and uses DenseNet121 \cite{huang2017densely} as the backbone to train a patch-to-spot regression model for predicting gene expression.

% Similarity-based method
Following this, mainstream methods have primarily focused on leveraging image similarity \cite{yang2023exemplar, xie2024spatially} and spatial relationships \cite{pang2021leveraging,zeng2022spatial} to enhance model performance. EGN \cite{yang2023exemplar} introduces an exemplar learning mechanism to dynamically select the most similar point within the WSI for a target point. 
In contrast, BLEEP \cite{xie2024spatially} utilizes a bi-modal embedding architecture to estimate gene expression at queried points by identifying similar expression profiles. 
However, variations in staining styles and specimen preparation can lead to significant differences at the gene expression level for points identified through image similarity. This limits generalization capability when models encounter entirely new, unseen samples.

% spatial information-based method
On the other hand, methods such as HisToGene \cite{pang2021leveraging} and His2ST\cite{zeng2022spatial} enhance performance by incorporating spatial relationships between spots within a single-layer WSI, utilizing Transformers and graph neural networks to capture global and neighborhood spatial information.
TRIPLEX \cite{chung2024accurate} introduces multi-resolution information centered on spots, integrating spatial information between patches and WSIs for prediction. 
However, these methods are limited to single-layer 2D spatial relationships and overlook the overall distribution of tissues in 3D space, thereby constraining a comprehensive understanding of spatial interactions.

% Our strength
In contrast, our proposed ASIGN framework leverages the integration of 3D-level information associations and sample-level training strategies to better capture the spatial distribution of information, thus significantly enhancing the generalization capability for cross-sample predictions.

\subsection{3D ST Analysis and Label Imputation}
% 3D Spatial Transcriptomic
As spatial transcriptomics technology has gradually developed in recent years, some studies have successfully constructed 3D spatial transcriptomic data for the mouse brain \cite{ortiz2020molecular} and cancer tissues \cite{andersson2021spatial}. Researchers conducted an in-depth analysis of 3D spatial transcriptomic data to construct a dynamic map of tissue changes across temporal and spatial dimensions, revealing the pathways of disease progression \cite{wu2021single} and the complex mechanisms of gene expression and spatial structural changes during organismal development \cite{maynard2021transcriptome, wang2022high}.

However, building up a 3D ST dataset is extremely costly. To narrow this gap, other studies have attempted to use this 3D spatial transcriptomic information to reconstruct and predict gene expression patterns. Existing 3D spatial transcriptomic methods primarily align different slice layers using gene-level expression similarities between spots \cite{zeira2022alignment, wang2023construction}. 
For example, PASTE \cite{zeira2022alignment} employs Gromov-Wasserstein optimal transport to probabilistically align neighboring slices based on transcriptional and spatial similarities. 
Stitch3D \cite{wang2023construction} uses a graph neural network to map gene expression and spatial data from multiple slices into a shared latent space to reconstruct 3D spatial transcriptomic. 
However, due to deformation and displacement during slicing, spot-level 3D reconstruction often fails to capture true spot distributions.
% Label Imputation
BLEEP \cite{xie2024spatially} proposes query-reference imputation to directly weight and predict labels for unknown spots, but information shifts between spots and questionable query reliability have limited its effectiveness.

% limitation
On the contrary, our approach performs registration at the WSI level to avoid spot-level deformation and displacement. We also optimize label information by combining deep learning with a cross-layer imputation block to improve label imputation accuracy.

\section{Method}
\subsection{Overall Framework}

Our proposed ASIGN architecture consists of three main components, including 3D graph construction, MSAGNet, and the CLI block. Figure \ref{fig:Overall_framework} shows the whole process of our ASIGN framework.
First, we align the original 2D single-layer WSIs using a global alignment method to estimate cross-layer overlap and similarity, thereby facilitating the construction of spot-level 3D graph connections. 
Then, we leverage MSAGNet to integrate multi-resolution image information and incorporate 3D spatial relationships, enhancing the model's representation capabilities. 
Finally, the CLI module propagates known labels to unknown spots, combining adaptive weights with model predictions to generate the final results.

\subsection{3D Graph Construction}

WSI-wise alignment is a critical prerequisite for obtaining 3D information distribution. First, we utilize XFeat \cite{potje2024xfeat} to extract features from each WSI sample and align them using Advanced Normalization Tools (ANTs) \cite{tustison2021antsx}. We take the middle layer of each sample as a reference for image registration.

After alignment, for the $j$-th spot in the $i$-th layer, denoted as $s_{ij}$, we draw a circle centered at its midpoint, denoted as $c_{ij}$. 
Subsequently, we calculate the spatial connection weight between this point and any spot $s_{pq}$ on other layers using Intersection over Union and cosine similarity, as follows:
\begin{equation}
e_{ij}^{pq} = \frac{|c_{ij} \cap c_{pq}|}{|c_{ij} \cup c_{pq}|} + \frac{\vec{c}_{ij} \cdot \vec{c}_{pq}}{\|\vec{c}_{ij}\| \times \|\vec{c}_{pq}\|}
\end{equation}

where $\vec{c}_{ij}$ is the extracted feature from ${c}_{ij}$. 
For spots $s_{ik}$ on the same layer$i$, we employ the Euclidean Distance to compute their connection weight, defined as $e_{ij} ^{ik}$. 
Finally, for each $s_{ij}$, we select the Top-k values of $e_{ij}^{pq}$ and $e_{ij}^{ik}$ to build up connections with spots other layers and current layer respectively. 
Then the constructed 3D graph is utilized in the training of MSAGNet.

\subsection{MASGNet}
Figure \ref{fig:network_structure} illustrates the network structure of MSAGNet and the CLI block. To effectively integrate multi-level information, MSAGNet leverages cross-attention layers to fuse features from multi-resolution patches and employs a GAT-Transformer block to aggregate sample-level 3D information. 
Subsequently, a regression head generates predictions for each resolution level independently.

\subsubsection{Hierarchical Fusion of Multi-Level Features}
For each spot \(s\), the paired images consist of spot-level, region-level, and global-level patches, denoted as $i_s$, $i_r$, and $i_g$ respectively. 
A pretrained ResNet50 \cite{he2016deep} is employed as the encoder for the feature extraction process to obtain features at different resolution levels, denoted as $f_s$, $f_r$, and $f_g$. 
To reduce computational costs, we follow the approach proposed by TRIPLEX \cite{chung2024accurate} and freeze the parameters of the encoders for $i_r$ and $i_g$, only updating the parameters of the spot-level encoder.

We employ a cross-attention layer to incorporate information from $f_r$ and $f_g$ into $f_s$, optimizing the representation of $f_s$ by aligning multi-source information. 
Taking the fusion with region-level information as an example, we treat $f_s$ as the query matrix and $f_r$ as the key and value matrices. The fused feature $f'_s$ is given by the following formula:
\begin{equation}
f'_s = \text{softmax} \left( \frac{f_s f_r^T}{\sqrt{d}} \right) f_r
\end{equation}
where $\sqrt{d}$ is a scaling factor based on the feature dimension. 
Similarly, we can obtain the fused feature $f''_s$ by aligning $f_s$ with global-level information.
Finally, we concatenate the features from the three levels to obtain the fused multi-level feature $F$, represented as:
\begin{equation}
F = \text{concat}(f_s, f'_s, f''_s)
\end{equation}

\subsubsection{Spatial-aware Graph Attention Block}
To integrate feature information in 3D space, we employ a Spatial-aware Graph Attention Block that combines multi-layer Graph Attention Networks and Transformer layers to achieve feature fusion.

Based on the 3D graph construction strategy proposed in Section 3.1, we establish edge connections for each resolution level, forming a graph structure in the following GAT block for processing. 
Taking the feature $F_i$ corresponding to spot $s_i$ as an example, after rounds of graph convolution, the processed feature $F'_i$ is expressed as follows:

\begin{equation}
\mathbf{F'}_i = \bigg\Vert_{k=1}^K \sigma \left( \sum_{j \in \mathcal{N}(i)} \alpha_{ij}^k \mathbf{W}^k \mathbf{F}_j \right)
\end{equation}

where $\mathcal{N}(i)$ denotes the set of neighboring nodes, $\bigg\Vert$ represents the concatenation operation of multi-head attention, $\sigma$ is the activation function, $\alpha_{ij}^k$ is the attention weight of the $k$-th attention head, and $\mathbf{W}^k$ is a linear transformation matrix determined by the connections between nodes. 

The outputs $F'_i$ from each round are concatenated and, together with features processed at the Global-level and Region-level, are fed into a Transformer layer to achieve flexible information aggregation. 
Finally, a regression head generates predictions for the global-level, region-level, and spot-level, denoted as $p_g$, $p_l$, and $p_s$ respectively.

\subsection{Cross-layer Imputation}
To further enhance the generalization capability of the 3D framework, we introduce a Cross-layer Imputation block designed to propagate known gene expression label information from one layer to unknown spots in other layers. 
For nodes with known labels, their labels remain fixed during the propagation process. For an unknown node $n_i$, the prediction result $p'_i$ after $t+1$ iterations can be given as:

\begin{equation}
{p'}_{i}^{(t+1)} = \frac{\sum_{j \in \mathcal{N}(i)} \alpha_j \cdot e_{ij} \cdot {p'}_{j}^{(t)}}{\sum_{j \in \mathcal{N}(i)} e_{ij}}
\end{equation}

where $\mathcal{N}(i)$ denotes the set of neighbors of node $i$, $\alpha_{j}$ is a learnable weight for node j, $e_{ij}$ represents the edge weight between node $i$ and its neighboring node $j$ as mentioned in Section 3.1, and ${p'}_{j}^{(t)}$ is the label distribution of neighboring node $j$ at the $t$-th iteration. The denominator $\sum_{j \in \mathcal{N}(i)} e_{ij}$ serves as a normalization factor.

Subsequently, we combine the model prediction results and the imputation results using an adaptive weight $\alpha$, yielding the final spot-level prediction result $p'_s$ given by:

\begin{equation}
p'_s = \alpha \cdot p_s + (1 - \alpha) \cdot p'_i
\end{equation}

\subsection{Loss Function}
We adopt a hybrid loss function to optimize the model's integration and learning of multi-level information. 
The overall loss function consists of multi-level prediction loss ($L_p$) and multi-level consistency loss ($L_c$). 
The multi-level prediction loss evaluates the discrepancy between the model's predictions and the true labels. 
At the spot level, we use Mean Squared Error (MSE) and Pearson Correlation Coefficient loss (PCC loss) to assess the model's prediction performance. 
For the region-level and global-level, we only use PCC loss ($P$) to estimate the model's prediction performance, thereby avoiding the introduction of extra noise. Therefore, $L_p$ can be described as:
\begin{equation}
L_p = MSE(p_s, y_s) + \sum_{i=s,r,g} \lambda_i \cdot P(p_i, y_i)
\end{equation}
where $s$, $r$, and $g$ represent spot-level, region-level, and global-level, respectively. $p_i$ and $y_i$ denote the model's predictions and labels, and $\lambda_i$ is a hyperparameter used to balance the PCC loss across resolutions.

We further constrain the predictions across different resolutions by a cross-resolution consistency loss. 
Since the gene expressions of patches at different resolutions share similar trends, we employ PCC loss to assess the differences between spot-level prediction and region-level as well as global-level predictions. $L_c$ is formulated as:
\begin{equation}
L_c = \lambda_1 \cdot P(p_s, p_r) + \lambda_2 \cdot P(p_s, p_g)
\end{equation}
Thus, the overall loss of the model $L$ is described as:
\begin{equation}
L = \gamma_1 \cdot L_p + \gamma_2\cdot L_c
\end{equation}
where $\gamma_1$ and $\gamma_2$ are hyperparameters used to balance the two losses and are set as 1 in the following experiments.

\section{Data and Experiments}

\textbf{Dataset and pre-processing procedure.}
We utilized three public datasets for fair cross-validation: two breast cancer datasets \cite{he2020integrating, andersson2021spatial} and one human dorsolateral prefrontal cortex (DLPFC) dataset \cite{maynard2021transcriptome}. 
The HER2-positive breast tumor dataset is referred to as HER2, while the other breast dataset proposed by ST-Net is denoted as ST-Data.
The HER2 dataset consists of 4 samples with 6-layer sections and 4 samples with 3-layer sections, and a total of 13,620 spots. The DLPFC dataset contains 3 samples with 4-layer sections and 42,474 spots. 
For optimal global alignment, we selected 16 three-layer samples from the 23 available samples in ST-Data, totaling 41,544 spots.

With over 20,000 original gene dimensions, predicting all gene expressions from histological image patches is impractical.
Therefore, we referred to the method mentioned in ST-Net \cite{he2020integrating} selecting the top 250 genes with the highest average expression for prediction. The normalization of gene expression values was performed following the strategy proposed by TRIPLEX \cite{chung2024accurate}, with a proportional normalization followed by a log transformation.

We cropped patches at different levels into sizes: spot-level, region-level, and global-level, with dimensions $224\times 224$, $512\times512$, and $1024\times 1024$ pixels, respectively. Spot-level patches were centered on the spot, while other patches used overlaps to find corresponding spot patches.

\noindent \textbf{Evaluation metrics.}
The Pearson correlation coefficient (PCC), mean squared error (MSE), and mean absolute error (MAE) is employed as the evaluation metrics to provide a comprehensive assessment of the models' performance in gene expression prediction tasks.

\noindent \textbf{Implementation.}
The experiments were conducted using two NVIDIA RTX A6000 GPU cards. We utilized an SGD optimizer configured with a momentum of 0.9 and a weight decay parameter set to $10^{-4}$. The initial learning rate, denoted as $lr_0$, was set to $10^{-4}$. A cosine decay strategy was applied to reduce the learning rate to 0.01 of its initial value during training. We employed a batch size of 256 for training and fine-tuned the hyperparameters $\lambda_1$, $\lambda_2$, $\lambda_s$, $\lambda_r$, and $\lambda_g$ in our hybrid loss function to values of 0.25, 0.25, 0.75, 0.5, and 0.5, respectively. We set the ratio of known labels as 0.2 in the CLI block to achieve the ideal performance. The top k selection for 3D graph construction is set as 8 for inter-layer spots and 12 for cross-layer spots.

\begin{table*}[t]
    \centering
    \caption{Quantitative comparisons in both 2D WSI-wise and 3D Sample-wise methods across different public datasets.}
    \resizebox{\textwidth}{!}{%
    \begin{tabular}{cccccccccccc}
        \hline
        \multirow{2}{*}{Dimension}  & \multirow{2}{*}{Model} & \multicolumn{3}{c}{HER2\cite{andersson2021spatial}} & \multicolumn{3}{c}{DLPFC\cite{maynard2021transcriptome}} & \multicolumn{3}{c}{ST-Data\cite{he2020integrating}} \\

        \cline{3-11}

        & & MSE & MAE & PCC & MSE & MAE & PCC & MSE & MAE & PCC \\
        \hline
        \multirow{6}{*}{2D WSI-wise} & ST-Net\cite{he2020integrating} & 0.663
 & 0.633 & 0.430& 0.496
 & 0.552& 0.738 & 0.804 & 0.705& 0.517\\

        & EGN \cite{yang2023exemplar} & 0.580 & 0.585 &0.442 & 0.401& 0.490 & 0.757& 0.674 & 0.644 & 0.554\\
        & HisToGene\cite{pang2021leveraging} & 0.669 & 0.644 & 0.432 & 0.348  & 0.472 & 0.753 & 0.575 &0.598 &0.552  \\
        & BLEEP \cite{xie2024spatially} & 0.743 & 0.665 & 0.340 &  0.304 &0.429 &0.737
 & 0.581 & 0.590 & 0.496
 \\
        & His2ST \cite{zeng2022spatial}& 0.591& 0.591 & 0.441 &0.409 &0.494 & 0.764
& 0.534 &0.568 &0.539\\
        & \textbf{ASIGN 2D (Ours)} & 0.579 &0.590 & 0.476 & \textbf{0.275} & \textbf{0.411} &0.766 & 0.528 &0.566 &0.552 \\
        \hline
        \hline

        \multirow{3}{*}{3D Sample-wise} & Overlap &0.461 & 0.532 & 0.625 & 0.292 & 0.422 & 0.749 & 0.447 & 0.525 & 0.663\\
        & Similarity & 0.449 & 0.517 &0.649 &0.278 & 0.412 & 0.762 &0.579 & 0.548 &0.696  \\
            & \textbf{ASIGN 3D (Ours)} & \textbf{0.363} &\textbf{0.473}   &  \textbf{0.693}  & 0.353 & 0.458  & \textbf{0.773} &\textbf{0.332} &\textbf{0.459}   &\textbf{ 0.739}  \\
        \hline
    \end{tabular} 
    }
        \label{table:comparison}
\end{table*}

\subsection{Experimental Setting}
\noindent \textbf{Sample-level cross-validation strategy}
We performed cross-validation at the sample level across the three datasets, utilizing four-fold cross-validation for HER2 and ST-Data, and three-fold cross-validation for DLPFC to account for dataset size differences. During validation, one layer was selected as the known label layer and included in the training set, while other unknown layers were reserved for the test set. This approach ensures that all models have equitable access to known information, thereby maintaining a consistent and fair evaluation framework and allowing for a comprehensive assessment of model performance across varying data distributions.

\noindent \textbf{Baseline methods}
We benchmarked our model's performance against current SOTA methods, including 1) 2D-level CNN-based models such as ST-Net \cite{he2020integrating}, EGN \cite{yang2023exemplar}, and BLEEP \cite{xie2024spatially}, and 2) 2D-level Transformer and GNN-based networks like HisToGene\cite{pang2021leveraging} and His2ST\cite{zeng2022spatial}. Additionally, we designed and introduced two 3D-based imputation methods inspired by PASTE \cite{zeira2022alignment}, which utilize spot overlap information and patch similarity to enhance predictions. All baseline models were trained and evaluated under consistent conditions for fair comparison.

\section{Results}
\subsection{Empirical Validation}

The cross-validation performance of different baseline models across three datasets is summarized in Table \ref{table:comparison}. Our proposed ASIGN framework consistently outperformed existing approaches in terms of MSE, MAE, and PCC metrics across both the 2D WSI level and the 3D sample level experiments, with particularly strong results observed for the ASIGN 3D model. For instance, in the HER2 dataset, known for its high inter-sample heterogeneity, the ASIGN 3D model achieved MSE, MAE, and PCC values of 0.363$\pm$0.054, 0.473$\pm$0.053, and 0.693$\pm$0.091, respectively. This marks a substantial improvement over other methods, such as the 3D similarity-based approach (PCC = 0.649$\pm$0.130) and the 2D EGN model (PCC = 0.442$\pm$0.152). These results indicate that the ASIGN framework is particularly adept at capturing complex data relationships that are critical for accurate modeling in heterogeneous datasets.

\begin{figure*}[htb]
    \centering
    \includegraphics[width=\linewidth]{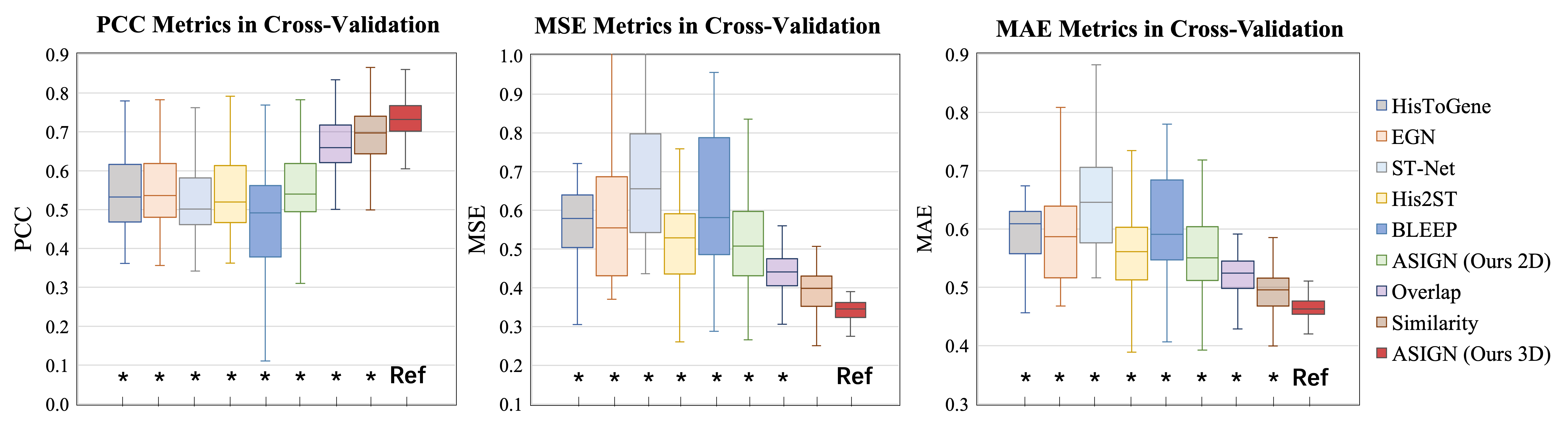}
\caption{\textbf{Quantitative comparison between different methods.} This figure presents the box plot of different methods with Pearson correlation coefficient (PCC), mean squared error (MSE), and mean absolute error (MAE). Welch’s t-test is used to assess the statistical significance of differences between other method and ASIGN. P-values below 0.05 are considered significant and are annotated with $\ast$.}
    \label{fig:boxplot}
\end{figure*}

Moreover, the boxplot presented in Figure \ref{fig:boxplot} highlights the generalization capability and prediction stability of the models under cross-dataset validation. Our ASIGN architecture exhibited the smallest prediction deviation across different datasets, underscoring its robust cross-sample prediction generalization capabilities. As it ensures consistent performance across diverse patient data, further enhancing the framework's practical value.

This significant improvement stems from the ASIGN framework's exceptional ability to accurately capture and integrate multi-dimensional information through the construction of cross-layer connections, rather than focusing solely on intra-layer information as is typical with existing methods. The ASIGN framework also facilitates the effective propagation of known layer information into unexplored regions, thereby enhancing the model’s generalization ability and providing a more applicable approach to clinical uses.

\subsection{Cancer Marker Gene Expression Prediction}
We identified two key cancer markers, Human Epidermal Growth Factor Receptor 2 (ERBB2) \cite{mehta2014co, dent2013her2} and Midkine (MDK)\cite{aynaciouglu2019involvement, zhang2021midkine}, which are highly associated with HER2 and have significant clinical relevance, to evaluate the model's applicability in real clinical scenarios. 

\begin{figure*}[htb]
    \centering
    \includegraphics[width=\linewidth]{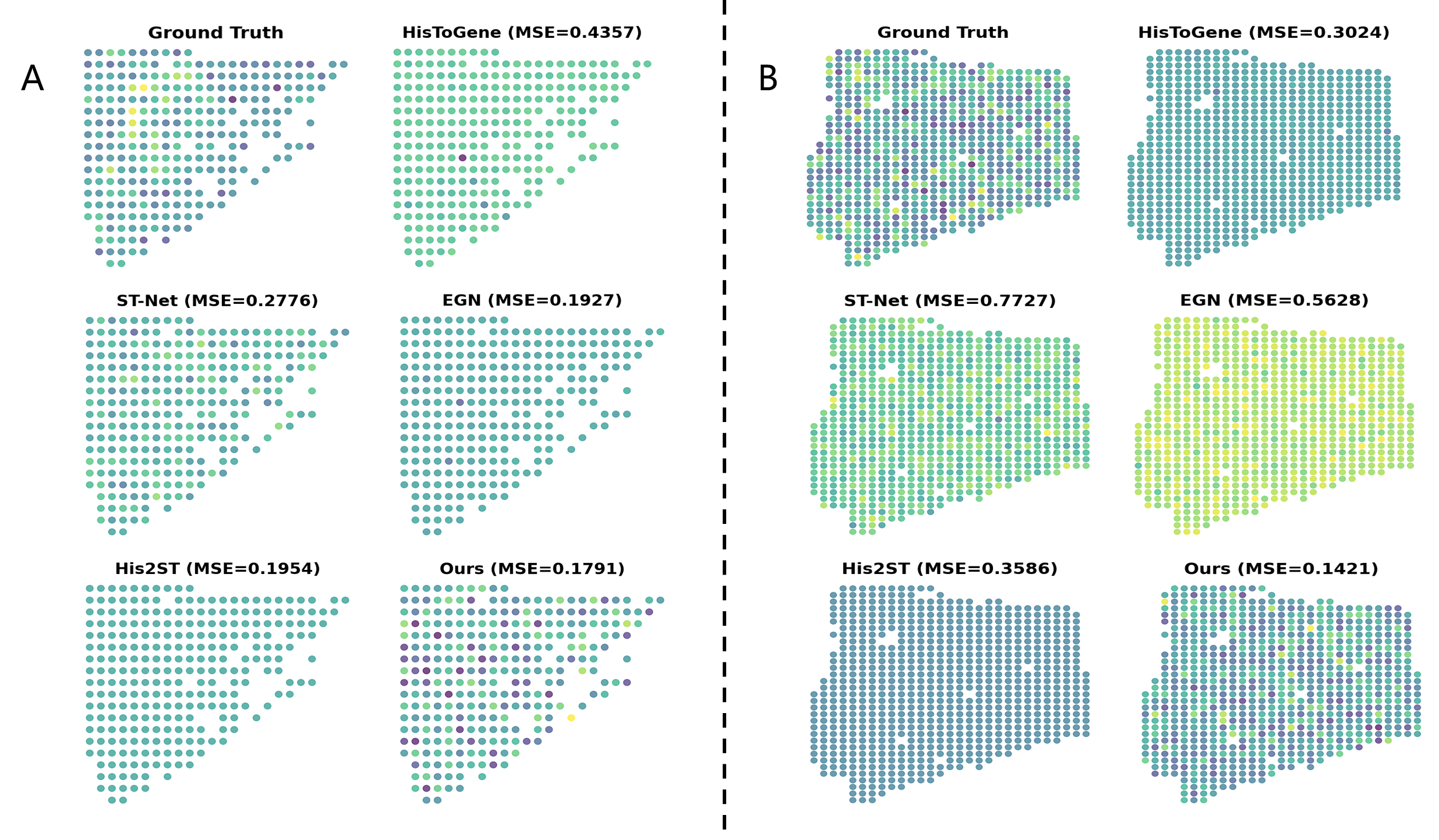}
    \caption{\textbf{Qualitative comparison between different methods.} This figure presents the ST imputation performance for predicting the distribution of cancer markers, specifically (A) ERBB2 and (B) MDK, within the WSIs. The MSE between the ground truth and the predicted values is presented for each model.}
    \label{fig:biomarker}
\end{figure*}

Figure \ref{fig:biomarker} visualizes the spatial distribution of ERBB2 and MDK in WSIs, along with the MSE metrics of different models for these critical genes. Our model achieved the lowest MSE for ERBB2 and MDK, at 0.1421 and 0.1791, respectively. In contrast, some 2D-based methods, such as HisToGene (0.8200) and His2ST (0.7410), failed to effectively predict these key cancer markers due to significant variance in cross-sample prediction, underscoring their limitations in capturing cross-sample spatial heterogeneity. By comparison, our ASIGN framework was able to nearly replicate the gene expression spatial distribution of these key markers, demonstrating its robustness and superior capacity to capture complex patterns. This makes it more applicable for pathologists to use in clinical practice and further enhances its practical utility in real-world scenarios.

\subsection{Ablation Study}

We designed three sets of ablation experiments to verify the effectiveness and necessity of each module within the ASIGN framework. These experiments include 1$)$ the comparison between 2D-based and 3D-based graph construction 2$)$ the network functional block, and 3$)$ the CLI block and the proportion of known labels

We adhered to the cross-validation procedure employed in the comparison experiments to ensure a fair evaluation of the ablation study. We trained the models on the HER2 dataset using identical hyperparameter settings, which allowed us to objectively assess the impact of each module and isolate their individual contributions to the overall performance of ASIGN.

\noindent \textbf{Function blocks in ASIGN}  
The experimental results for the ablation study of the functional blocks incorporated in ASIGN are shown in Table \ref{table:blocks}. The cross-attention layer for fusing multi-resolution features and the GAT blocks for capturing 3D spatial information improved the PCC metric from 0.659 to 0.677 and 0.686, respectively. Moreover, when both blocks were utilized together, the model's performance further increased to 0.693. Notably, the benefits from the two blocks are orthogonal, enabling ASIGN to achieve optimal performance when both are employed together.

This indicates that the cross-attention layer plays a crucial role in effectively integrating 2D multi-resolution information, enabling the model to capture and align details at different scales. Meanwhile, the GAT blocks are responsible for processing and handling 3D spatial information, providing an additional layer of contextual understanding through 3D spatial relationships. By strategically combining these multi-level and multi-source features, the model can develop a more holistic data representation.  

\begin{table}[ht]
    \centering
    \caption{Ablation study for functional blocks in ASIGN}
    \resizebox{0.45\textwidth}{!}{
    \begin{tabular}{lccc}
        \hline
        Functional Blocks & MSE  & MAE  & PCC \\
        \hline
        w.o GAT \& Cross-atten & 0.674 & 0.531 & 0.659 \\
        w.o Cross-atten & 0.578 & 0.692 & 0.686  \\
        w.o GAT & 0.466 & 0.627 &  0.677  \\
        w. Both blocks & \textbf{0.363} & \textbf{0.473} & \textbf{0.693}  \\
        \hline
    \end{tabular}
    }
    \label{table:blocks}
\end{table}

\noindent \textbf{CLI block and proportion for known labels.}  
Table \ref{table:CLI} illustrates the impact of different proportions of known labels on the model's predictive performance. When the CLI block is introduced to propagate known labels to unknown labels, the model's performance sees a significant improvement, with the PCC metric increasing from 0.635 to 0.693. This underscores the essential role of the CLI block and label imputation in enhancing the effectiveness of our proposed ASIGN framework. 

Additionally, as the proportion of known labels continues to increase, the model's performance progressively improves. Once the proportion of known labels reaches the range of 20-30\%, the model performance stabilizes, indicating convergence. This suggests that beyond this threshold, additional known labels provide diminishing returns, affirming that a relatively small proportion of known labels can yield optimal predictive outcomes within our proposed ASIGN framework.

\begin{table}[ht]
    \centering
    \caption{Ablation study for CLI block and known label proportion}
    \resizebox{0.45\textwidth}{!}{
    \begin{tabular}{lccc}
        \hline Proportion of Known Label
        & MSE  & MAE  & PCC\\
        \hline 
        w.  0\%  (w.o CLI)  & 0.622 & 0.606 & 0.635 \\
        w. 10\%   &  0.542 & 0.563  & 0.677  \\
        w. 20\%  (ASIGN) &  \textbf{0.363} & \textbf{0.473} & 0.693  \\
        w. 30\%  & 0.397 &  0.492& \textbf{0.708}  \\
        \hline 
    \end{tabular}
    }
    \label{table:CLI}
\end{table}

\noindent \textbf{Comparison of 2D and 3D graph construction.}  
Table \ref{table:dimension} compares the performance of 2D-based and 3D-based graph construction on the HER2 dataset. After incorporating 3D spatial information, the model's PCC score significantly increased from 0.476 to 0.635. This demonstrates that adding an extra spatial dimension allows the model to better capture gene correlations using 3D spatial information and reconstruct the true three-dimensional structure of tissues. Consequently, this wealth of comprehensive information allows for a more detailed and precise analysis of gene expression patterns, significantly enhancing the model’s predictive capability and overall accuracy, thus making reliable predictions across diverse datasets.

\begin{table}[ht]
    \centering
    \caption{Ablation study for graph construction}
        \resizebox{0.45\textwidth}{!}{
    \begin{tabular}{cccc}
        \hline Graph Construction
        & MSE  & MAE  & PCC\\
        \hline
        2D WSI-wise &  \textbf{0.579} &\textbf{0.590} & 0.476 \\ 
        3D Sample-wise &  0.622 & 0.606 & \textbf{0.635} \\
        \hline
    \end{tabular}
    }
    \label{table:dimension}
\end{table}

\section{Conclusion}
We propose a new learning paradigm for 3D ST modeling that imputes full 3D volumetric ST data using 3D WSI sections combined with a single 2D ST acquisition. This approach provides a balanced solution—more precise than “free but less accurate” 3D predictions, yet far more affordable than the “precise but prohibitively expensive” acquisition of full 3D ST for every slide. To this end, our ASIGN framework integrates multi-level features, combining local multi-resolution patches with global 3D sample-wise graph connections to enhance prediction and imputation outcomes. Experimental results demonstrate ASIGN's superior performance and robust generalization in cross-sample validation across multiple public datasets, significantly improving the PCC metric for gene expression tasks from $\approx$0.5 with existing methods to $\approx$0.7. ASIGN offers a promising pathway to achieve accurate and cost-effective 3D ST data for real-world clinical applications.
\section*{Acknowledgment}     
This research was supported by NIH R01DK135597(Huo), DoD HT9425-23-1-0003(HCY), NIH NIDDK DK56942 (ABF). This work was also supported by Vanderbilt Seed Success Grant, Vanderbilt Discovery Grant, and VISE Seed Grant. This project was supported by The Leona M. and Harry B. Helmsley Charitable Trust grant G-1903-03793 and G-2103-05128. This research was also supported by NIH grants R01EB033385, R01DK132338, REB017230, R01MH125931, and NSF 2040462. We extend gratitude to NVIDIA for their support by means of the NVIDIA hardware grant. 

{
    \small
    \bibliographystyle{IEEEtran}
    \bibliography{main}
}

\clearpage
\setcounter{page}{1}

% \section*{Supplementary Material}
\appendix
\label{sec:Supplementary Material}

\subsection{Details in Public Datasets} 
We employed the HER2~\cite{andersson2021spatial}, ST-data~\cite{he2020integrating}, and DLPFC~\cite{maynard2021transcriptome} datasets for cross-validation of our model. The ST-data dataset comprises 23 breast cancer histopathological samples, each consisting of 2 or 3 adjacent layers with a thickness of 16~$\mu$m. The HER2 dataset includes 8 samples containing 6 or 3 adjacent layers, while the DLPFC dataset features 3 samples, each composed of 4 directly adjacent 10~$\mu$m consecutive tissue sections from the human dorsolateral prefrontal cortex. All whole-slide images (WSIs) were scanned at 10x magnification.

To ensure the accuracy of 3D sample-level information, we excluded 7 samples from the ST-data dataset due to suboptimal registration quality. Figure~\ref{fig:profile} provides detailed information about each sample used in the three datasets, along with their summarized profiles.

\subsection{Multi-resolution Patches and Gene Selection}
\begin{figure*}[htb]
    \centering
    \includegraphics[width=\linewidth]{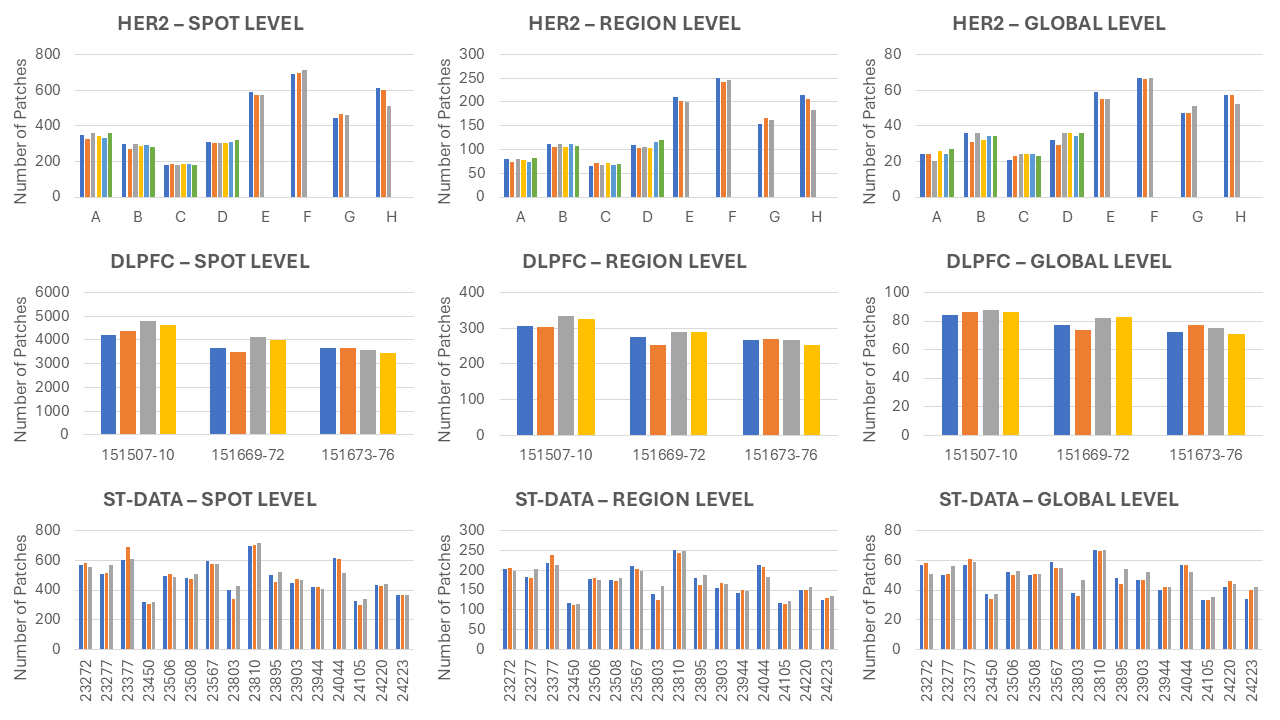}
    \caption{\textbf{Profile of public datasets.} This figure provides an overview of the profiles and the distribution of multi-resolution patches extracted from the HER2, DLPFC, and ST-Data datasets across multi-layer sample levels. }
    \label{fig:profile}
\end{figure*}

We extracted patches at multiple resolutions, specifically at the spot level, region level, and global level, by cropping the images to sizes of $224 \times 224$, $512 \times 512$, and $1024 \times 1024$ pixels, respectively. The relationships between patches across different resolution levels were established by analyzing their spatial overlaps, ensuring consistent alignment across scales. Detailed information on the multi-resolution patches, including their distribution and characteristics for each public dataset, is presented in Figure~\ref{fig:profile}.

To determine the ground truth values for the region and global levels, $y_r$ and $y_g$, we aggregated the spot-level gene expression values within their respective region and global areas. This process is mathematically expressed as:

\begin{equation} y_r = \sum_{i \in R} y_i, \quad y_g = \sum_{i \in G} y_i \end{equation}

Here, $y_i$ denotes the gene expression value at the $i$-th spot, $R$ represents the set of spots within a specific region, and $G$ denotes the set of spots within the global area. This approach ensures a consistent representation of gene expression across multiple spatial levels.

\begin{figure*}[htb]
    \centering
    \includegraphics[width=\linewidth]{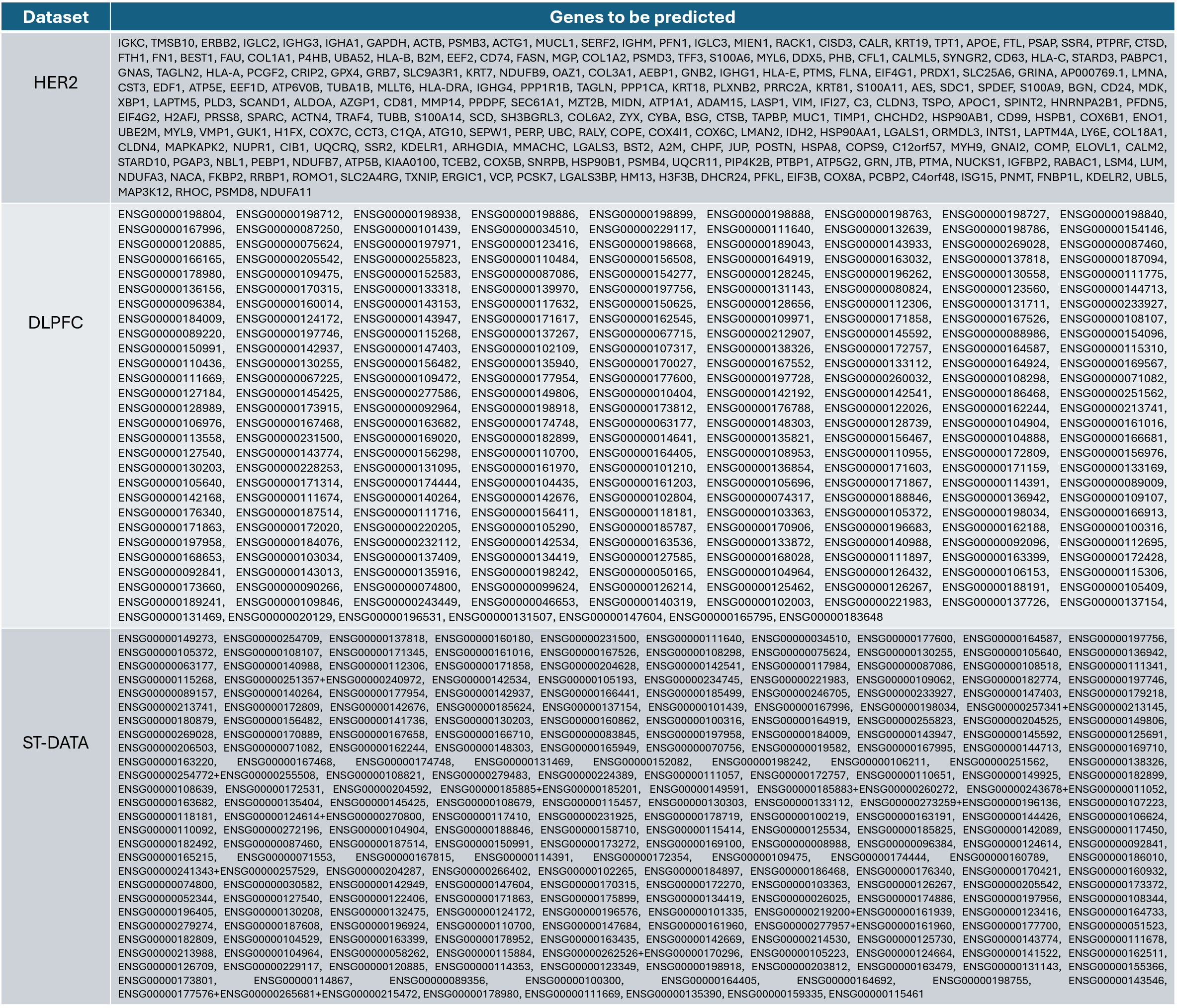}
    \caption{\textbf{Genes selection in each public dataset.} This figure showcases the top 250 genes with the highest expression levels or their corresponding codes for each public dataset utilized in this task.}
    \label{fig:selected_gene}
\end{figure*}

Following the methodology outlined in ST-Net~\cite{he2020integrating}, we selected the top 250 genes with the highest average expression for prediction. Figure~\ref{fig:selected_gene} provides details of the selected genes and the corresponding codes for each dataset.

\subsection{Data Augmentation}

To improve the model's generalization ability and ensure consistency across all experiments, we implemented a standardized data augmentation strategy. This approach included random horizontal and vertical flips to simulate variations in image orientation, random 90-degree rotations to enhance rotational invariance, and final normalization to adjust pixel values to align with the standard distribution used by pretrained models (mean [0.485, 0.456, 0.406] and standard deviation [0.229, 0.224, 0.225]).

\subsection{3D Global Alignment}

\begin{figure*}[htb]
    \centering
    \includegraphics[width=\linewidth]{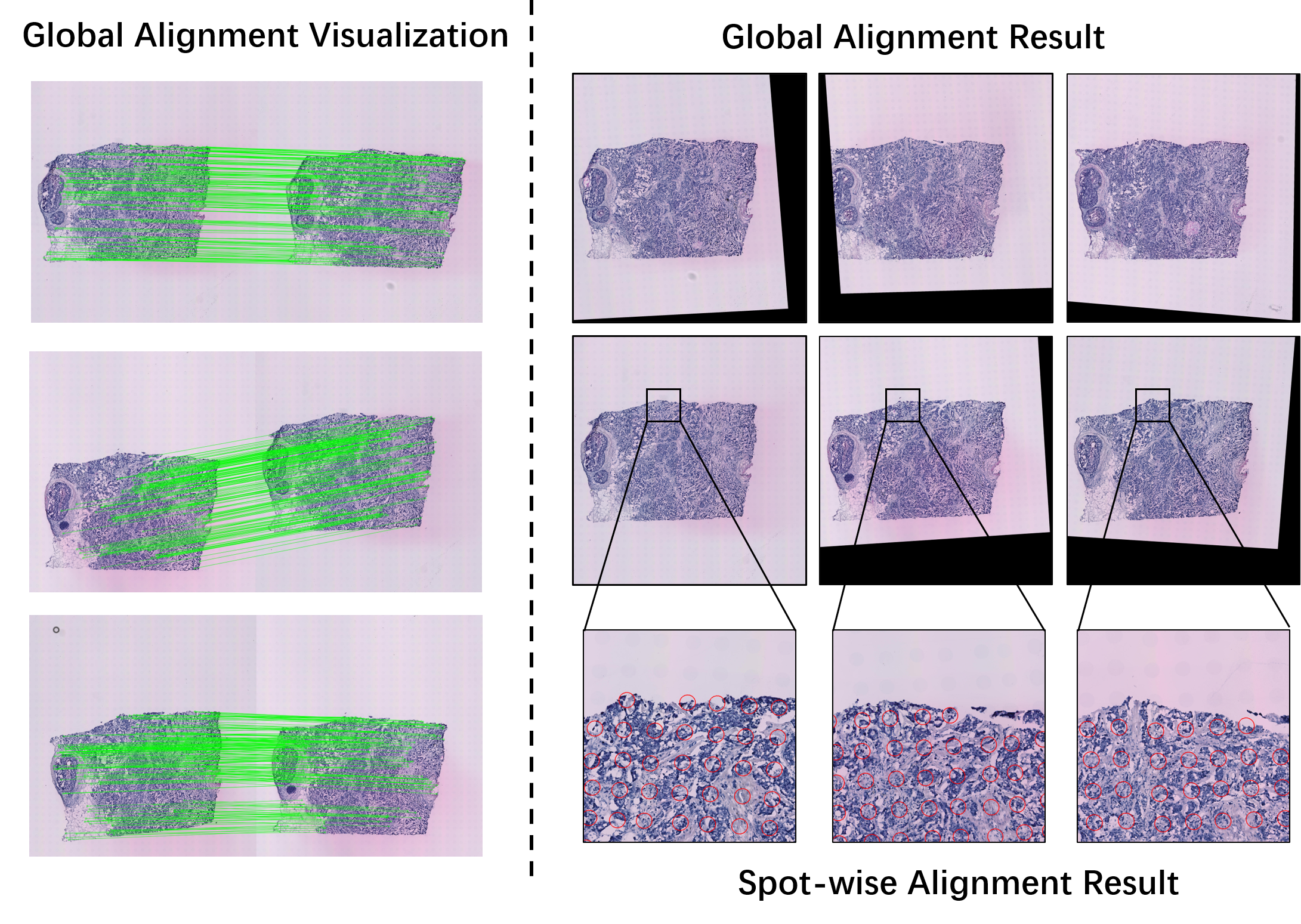}
    \caption{\textbf{Visualization of 3D global alignment.} This figure illustrates the 3D global alignment results for a 6-layer sample, showcasing both WSI-level and spot-level alignments. The red circles on the patches represent the spots following 3D global alignment.}
    \label{fig:alignment}
\end{figure*}

We utilized Xfeat~\cite{potje2024xfeat} for feature extraction and ANTs~\cite{tustison2021antsx} for WSI-level image registration. Figure~\ref{fig:alignment} visualizes the 3D global alignment results and the aligned spots for a 6-layer sample. To achieve this, we extracted geometric transformation parameters between images (scaling, rotation, and translation), generated inverse affine matrices for each image relative to the reference image, and combined multiple transformation matrices through matrix multiplication to ensure effective transmission of registration information across different layers. 

The middle image was designated as the reference image and left untransformed, serving as the baseline, and directly saved to the reconstruction directory to minimize accumulated computational errors. During the registration process, transformations were applied recursively: forward for images preceding the middle image and backward for those following it, completing a bidirectional registration workflow. To accelerate the process, the symmetric normalization algorithm was employed using 8 threads.

Following the completion of the 3D WSI global alignment, we performed the 3D spot-level alignment. For each spot in the original image, circles with a radius of 112 pixels were drawn, and 100 points were plotted within each circle. Using the transformation matrices derived from the global alignment, these circles were aligned at the spot level to identify cross-layer spot overlap information. This overlap data was subsequently utilized to construct the 3D sample-level graph by establishing connections between spots.

\subsection{Implementation Details for Baselines}

To ensure a fair and consistent comparison between the baseline methods and our proposed ASIGN framework, we made minimal yet necessary adjustments to the implementation of certain baseline models. These modifications were carefully designed to align input sizes, feature extraction strategies, and imputation methods across all models, ensuring that performance differences could be attributed solely to the respective methodologies rather than inconsistencies in the experimental setup.

For \textbf{HisToGene}~\cite{pang2021leveraging} and \textbf{His2ST}~\cite{zeng2022spatial}, which originally utilize spot images of $112 \times 112$ pixels, we resized all images to a fixed size of $224 \times 224$ pixels to ensure consistent input dimensions across models.

For \textbf{EGN}~\cite{yang2023exemplar}, we replaced the originally used GAN-based encoder with a pretrained ResNet50 as the feature extractor, ensuring a consistent feature extraction strategy across all baseline models. In line with the original EGN implementation, we selected the 8 most similar exemplars for each spot for training.

For \textbf{BLEEP}~\cite{xie2024spatially}, we utilized the CLIP model as the backbone and applied the simple average strategy proposed by BLEEP to compute the mean gene expression values of the top 50 nearest spots, which served as the imputation prediction result.

For the \textbf{3D Sample-level Similarity-based Imputation}, we calculated the mean gene expression values of the top 20 nearest spots for each spot. This approach ensured a fair comparison with our proposed ASIGN framework by maintaining the same number of reference spots for imputation.

Additionally, we applied the data augmentation strategy described in the previous section uniformly across all models to ensure consistency in data preprocessing. During training, each model was trained until convergence, which was defined as the point where the loss stabilized and no longer exhibited significant changes over successive iterations.

\subsection{Implementation Details for ASIGN}
In our implementation of the proposed ASIGN framework, we utilized a pretrained ResNet50 to extract 1024-dimensional features from patches at multiple resolutions. A 3-layer GAT block was employed, with the output of each layer passed through a corresponding transformer layer. The features refined by each transformer layer were then weighted and aggregated to generate the final feature representation, which was subsequently fed into a linear layer to produce the model's prediction.

For each spot, we selected the top 12 nearest spots to establish edges and construct a 2D WSI-level graph. The edge weights were defined as the inverse of the Euclidean distance between the centers of the connected spots.

In 3D partially-known scenarios, we designated the first layer of each sample as the known layer for the experiments. For the label propagation block, we performed 10 iterations to ensure that all unknown nodes were adequately covered while avoiding over-propagation, which could lead to label homogenization and diminish prediction accuracy. Finally, an adaptive weighting mechanism was employed to combine the model's prediction results with the imputation results, producing the final spot-level prediction outputs that effectively leverage both direct predictions and inferred information.

\begin{table}[ht]
    \centering
    \caption{Quantitative comparisons between ASIGN and TRIPLEX}
        \resizebox{0.45\textwidth}{!}{
    \begin{tabular}{ccccc}
        \hline Dataset & Model
        & MSE  & MAE  & PCC\\
        \hline
        \multirow{3}{*}{HER2 ~\cite{andersson2021spatial}} & TRIPLEX~\cite{chung2024accurate}  & 0.592 & 0.591 & 0.450 \\
        & ASIGN 2D  &  0.579 & 0.590  & 0.476  \\
        &  ASIGN 3D  &  \textbf{0.363} & \textbf{0.473} &  \textbf{0.693}  \\
         \hline
         \hline
        \multirow{3}{*}{DLPFC~\cite{maynard2021transcriptome}} & TRIPLEX~\cite{chung2024accurate}  & 0.645 & 0.621 & 0.766 \\
        & ASIGN 2D  &  \textbf{0.275} &\textbf{0.411}  & 0.766  \\
        &  ASIGN 3D  &  0.353
        & 0.458 &  \textbf{0.773}  \\
        \hline
        \hline
        \multirow{3}{*}{ST-Data~\cite{he2020integrating}} & TRIPLEX~\cite{chung2024accurate}  & 0.627 & 0.621  &  0.553 \\
        & ASIGN 2D   &  0.528 &0.566  & 0.552  \\
        &  ASIGN 3D  &  \textbf{0.332} & \textbf{0.459} &  \textbf{0.739}  \\
        \hline
    \end{tabular}
    }
    \label{table:com_tri}
\end{table}

\subsection{Comparison with TRIPLEX}
\textbf{TRIPLEX}~\cite{chung2024accurate} is a recently proposed method that effectively utilizes multi-resolution information to generate accurate gene expression predictions. However, due to space constraints, a detailed comparison of TRIPLEX with our proposed ASIGN framework is not able to be included in the main text. 

In this section, Table~\ref{table:com_tri} provides a comprehensive comparison of these methods. The experimental results, evaluated across three key metrics, consistently demonstrate the superior performance of our 3D ASIGN framework. By leveraging 3D sample-level information and cross-layer label imputation strategy, ASIGN achieves significant improvements in all three metrics and outperforms all other baseline methods consistently.

\subsection{Additional Ablation Study on Public Datasets}

In this section, we provided the experimental results of ablation studies on the ST-Data and DLPFC datasets, which are not included in the main text due to space limitations. These experiments assess the contribution of each functional module in our proposed 3D ASIGN framework and examine the effects of varying known label proportions on gene expression prediction. The results, shown in Table~\ref{table:ablation_dorsal} and Table~\ref{table:ablation_ST}, reveal a consistent trend across all datasets, mirroring the findings from the HER2 dataset. This consistency underscores the complementary benefits of each functional block, enabling ASIGN to deliver optimal performance when all modules are integrated.

\begin{table}[ht]
    \centering
    \caption{Ablation study on DLPFC Dataset}
    \resizebox{0.45\textwidth}{!}{
    \begin{tabular}{lccc}
        \hline Ablation Study
        & MSE  & MAE  & PCC\\
        \hline
        w.o GAT \& Cross-atten & 0.405 & 0.497 & 0.767 \\
        w.o Cross-atten & 0.364 & 0.470 &  0.770  \\
        w.o GAT & 0.368& 0.475 & 0.768  \\
        w. Both blocks & \textbf{0.353} & \textbf{0.458} & \textbf{0.773}  \\
        \hline
        \hline
         w.  0\%  (w.o CLI)  & 0.420 & 0.503 & 0.770 \\
        w. 10\%   &  0.402 & 0.494  & 0.772  \\
        w. 20\%  (ASIGN) &  0.353 & 0.458 & \textbf{0.773}  \\
        w. 30\%  & \textbf{0.278}& \textbf{0.416}  & 0.767 \\
        \hline
        \hline
         2D WSI-wise &  \textbf{0.275} &\textbf{0.411} & 0.766 \\ 
        3D Sample-wise &  0.420 & 0.503 & \textbf{0.770} \\
        \hline
    \end{tabular}
    }
    \label{table:ablation_dorsal}
\end{table}

\begin{table}[ht]
    \centering
    \caption{Ablation study on ST-Data}
    \resizebox{0.45\textwidth}{!}{
    \begin{tabular}{lccc}
        \hline Ablation Study
        & MSE  & MAE  & PCC\\
        \hline
        w.o GAT \& Cross-atten & 0.541 & 0.571 & 0.706\\
        w.o Cross-atten & 0.472 & 0.535 & 0.710   \\
        w.o GAT & 0.451 &  0.528 & 0.708    \\
        w. Both blocks & \textbf{0.332} & \textbf{0.459} & \textbf{0.739}  \\
        \hline
        \hline
         w.  0\%  (w.o CLI)  &0.485  &  0.545 &0.604 \\
        w. 10\%   & 0.406  & 0.499    & 0.708    \\
        w. 20\%  (ASIGN) & 0.332 & 0.459 & 0.739  \\
        w. 30\%  & \textbf{0.329} & \textbf{0.458}   &  \textbf{0.741} \\
        \hline
        \hline
         2D WSI-wise &  0.528 & 0.566 & 0.552 \\ 
        3D Sample-wise & \textbf{0.485}  & \textbf{0.545} & \textbf{0.604}  \\
        \hline
    \end{tabular}
    }
    \label{table:ablation_ST}
\end{table}

\end{document}